\documentclass{article}
\usepackage{arxiv}
\usepackage{hyperref}
\usepackage{url}
\usepackage{booktabs}
\usepackage{amsfonts}
\usepackage{nicefrac}
\usepackage{microtype}
\usepackage{graphicx}
\usepackage{natbib}
\usepackage{doi}
\usepackage{tabularx}
\usepackage{amsmath} 
\usepackage{multirow}

\title{DZIRI VoiceBOT: An End-to-End Low-Resource Speech-to-Speech Conversational System for Algerian Dialect}

\author{
(1) Dihia LANASRI, (2) Rebeh Imane AMMAR AOUCHICHE , (2) Abdelkarim REMMIDE,\\
\textbf{(2) Fairouz TAKI, (2) Asma KEMMOUM}\\
(1) ATM Mobilis, (2) Saad Dahlab Blida 1 University\\
Algiers, Algeria\\[0.5em]
\texttt{dihia.lanasri@gmail.com, imeneazouz@yahoo.fr,}\\
\texttt{abdelkarimremmide@gmail.com, fairouztaki19@gmail.com,}\\
\texttt{asmakemmoum09@gmail.com}\\
    \\    
}

\renewcommand{\shorttitle}{Dziri Voice Bot}

\hypersetup{
pdftitle={Dziri Voice Bot},
pdfsubject={NLP, Speech, DIALECT},
pdfauthor={D.LANASRI and al},
pdfkeywords={Natural Language Processing, Voice Bot, Algerian Dialect, Speech-to-Text, Text-to-Speech, ASR, RAG},
}

\begin{document}
\maketitle

\begin{abstract}
Automatic speech and language technologies are still heavily biased toward high-resource languages, limiting their applicability to dialectal and low-resource settings such as Algerian Dialect. This language presents additional challenges including lack of standardized orthography, frequent code-switching with French, and scarcity of annotated speech resources.

This paper addresses the problem of building a complete speech-to-speech conversational system for Algerian Dialect. We propose a modular pipeline integrating automatic speech recognition, natural language understanding, retrieval-augmented generation, and text-to-speech synthesis within a unified architecture.

This work is the continuation of our previous work on Algerian dialectal conversational systems \cite{bechiri2026dziribot}, extending it from text-based dialogue modeling to full speech-based interaction.

We constructed dedicated datasets for ASR, NLU, and TTS in the telecom domain and fine-tune pretrained models for each component. The ASR system is built on Whisper-based adaptation, while the NLU module combines transformer-based embeddings with a task-oriented dialogue framework. A neural TTS system is trained on a newly collected dialectal corpus to enable spoken response generation.

Experimental results show strong performance across all components, including low word error rate for ASR, high intent classification and entity recognition scores for NLU, and stable speech synthesis quality. The proposed system provides a reproducible baseline for end-to-end conversational modeling in Algerian Dialect.
\end{abstract}


\section{Introduction}
Recent progress in speech and language processing has led to the widespread deployment of conversational systems capable of handling spoken interactions across a variety of domains. Modern pipelines typically integrate automatic speech recognition (ASR), natural language understanding (NLU), dialogue management, and text-to-speech (TTS) synthesis into unified systems that enable natural human–machine communication. Despite these advances, the performance and usability of such systems remain strongly conditioned by the availability of large-scale annotated datasets and standardized linguistic resources.

This reliance on data abundance introduces a fundamental limitation: speech technologies are predominantly optimized for high-resource languages such as English, and a limited set of widely spoken languages. In contrast, a large number of the world’s languages and dialects remain underrepresented in both academic research and industrial applications. These low-resource and dialectal settings are characterized by limited or non-existent corpora, high linguistic variability, and the absence of standardized orthography, all of which significantly hinder the development of robust speech systems.

Algerian Dialect (Darija) exemplifies these challenges in a particularly acute form. It is a widely used spoken dialect across Algeria, estimated to be spoken by more than 45 million people in daily communication. However, it lacks formal standardization and is not supported by commercial speech technologies. Algerian Dialect exhibits substantial phonetic and lexical variability, frequent code-switching with French, and multiple writing conventions including Arabic script and Latin script (Arabizi). These properties make both acoustic modeling and language understanding significantly more complex than in standardized linguistic settings. As a result, users are often forced to interact with existing systems using Modern Standard Arabic or French, which do not reflect natural spoken usage and significantly reduce accessibility and adoption.

While multilingual and cross-lingual models have recently improved performance in low-resource scenarios, they still struggle to capture the full complexity of dialectal speech, particularly in task-oriented and domain-specific environments. In parallel, most existing conversational systems for Algerian dialects have been limited to text-based interaction \cite{boulesnane2022dzchatbot}. Although these systems demonstrate that task-oriented dialogue in Algerian Dialect is feasible, they do not address the full speech processing pipeline required for real-world voice-based applications, where both spoken input and spoken output are essential.

In this context, extending conversational systems from text-only interaction to full speech-to-speech pipelines becomes a necessary step toward practical deployment. Such systems must jointly handle noisy and highly variable speech input, perform robust semantic interpretation under code-switching conditions, and generate natural spoken responses in a dialect that lacks standardized phonetic and linguistic resources. This combination of challenges places Algerian Dialect among the most demanding scenarios for end-to-end spoken conversational AI.

This work is built upon prior research on Algerian dialectal conversational modeling \cite{bechiri2026dziribot}, which introduced a task-oriented text-based chatbot for Algerian Dialect. While that system demonstrated the viability of structured dialogue modeling in this dialect, it remained restricted to textual interaction and did not incorporate speech processing capabilities. The present work extends this line of research by introducing a complete speech-to-speech conversational framework that integrates ASR, NLU, retrieval-augmented generation (RAG), and TTS into a unified architecture designed specifically for Algerian Dialect.

To support this system, we constructed and curated new datasets covering speech recognition, intent classification, entity extraction, and speech synthesis in a domain-specific setting. Pretrained models are adapted and fine-tuned to address dialectal variability, code-switching, and low-resource constraints. The resulting system enables end-to-end spoken interaction in Algerian Dialect, bridging a gap between research prototypes and deployable conversational agents for under-resourced languages.

The proposed system is evaluated across all components of the end-to-end voice pipeline. First, we construct a dedicated Algerian Dialect speech corpus for the telecommunications domain (2.68 hours, 4,103 annotated utterances, 14 speakers, 70 intents) alongside a complementary NLU dataset containing 15,891 examples spanning 80 intents and 28 entities. Second, fine-tuning a Whisper-medium-based ASR model yields a word error rate of 13.74\%, representing the first reported ASR benchmark for Algerian Dialect in this domain.

For natural language understanding, a hybrid Rasa-based architecture enhanced with DziriBERT achieves 98.4\% intent classification accuracy and 93.9\% entity-level F1-score, outperforming prior reported results for dialectal Arabic in similar task-oriented settings. In addition, a conditional retrieval-augmented generation module is introduced to handle open-domain queries, achieving a 78.5/100 composite performance score while maintaining full technical reliability.

For speech synthesis, we build a dedicated Algerian Dialect TTS corpus (50.7 minutes, single speaker) and fine-tune two architectures, XTTS-v2 with LoRA adaptation and VITS-based models, establishing the first fine-tuned TTS systems for this dialect in a telecommunications context.

By addressing both linguistic and system-level challenges, this work contributes toward expanding the scope of speech-based artificial intelligence to dialectal environments that have so far remained largely unsupported in both academic literature and industrial deployment.

The remainder of this paper is organized as follows: Section 2 reviews related work in voicebots in Arabic and Algerian Dialect contexts, Section 3 presents the detailed approach that have been developed, Section 4 details the conducted experiments and obtained results, Section 5 presents a detailed discussion and analysis and Section 6 concluded the paper.

\section{Related Work}

The rapid development of speech and language technologies has enabled the emergence of advanced voice agents capable of supporting natural spoken interaction between humans and machines. These systems are now widely deployed in customer service, virtual assistants, and conversational AI platforms. However, their success is largely driven by the availability of large-scale annotated data, which makes them heavily biased toward high-resource languages. As a result, Arabic dialects, particularly North African varieties such as Algerian Dialect, remain significantly underrepresented in both academic research and industrial applications.

To situate our work, we review three main research directions relevant to end-to-end conversational systems: Automatic Speech Recognition (ASR), Natural Language Understanding (NLU), and Text-to-Speech (TTS) for Arabic dialects.

\subsection{Automatic Speech Recognition for Arabic Dialects}
Early ASR systems for Arabic relied on hybrid Hidden Markov Model (HMM) architectures combined with Weighted Finite State Transducers (WFST), as demonstrated by the ALASR system \cite{menacer2017development}, trained on Modern Standard Arabic (MSA) broadcast data. While such systems achieved reasonable performance in controlled settings, they struggle to generalize to dialectal speech due to lexical and phonetic variability.

Recent research has shifted toward end-to-end neural architectures and transfer learning from large pretrained models. For instance, \cite{ozyilmaz2025overcoming} investigated Whisper fine-tuning across multiple Arabic dialects, showing that dialect-pooled training can achieve performance comparable to dialect-specific models. Similarly, \cite{alblooki2025asr} evaluated pretrained models on under-resourced Arabic dialects and found that MMS-based systems perform strongly after adaptation, while Wav2Vec2 exhibits better zero-shot generalization.

For Algerian Dialect specifically, recent work has explored parameter-efficient adaptation techniques. Nasri (2024) proposed a Whisper-based model fine-tuned with LoRA adapters on a limited dialectal corpus, demonstrating the effectiveness of lightweight adaptation strategies for low-resource ASR. Earlier work on the ALGASD corpus introduced a classical HMM-based baseline system \cite{selouani2010algerian}, which was later extended using neural approaches such as Wav2Vec2 fine-tuning, achieving substantial reductions in Word Error Rate despite limited training data \cite{haboussi2025arabic}.

Multilingual and multi-task learning approaches have also been proposed to mitigate data scarcity. \cite{menacer2021investigating} demonstrated that combining Algerian Dialect, MSA, and French data within a joint training framework significantly improves recognition performance, highlighting the importance of cross-lingual transfer in low-resource settings.

Overall, existing ASR systems for Algerian Dialect remain either limited in scale or focused on isolated acoustic modeling, without integration into downstream conversational systems.

A key challenge in Algerian speech is pervasive code-switching between Arabic, French, and Arabizi, which significantly degrades the performance of standard ASR and NLU systems trained on monolingual data.

\subsection{Natural Language Understanding for Arabic Dialects}

Natural Language Understanding (NLU) for Arabic has benefited significantly from transformer-based language models. AraBERT \cite{antoun2020arabert} established strong performance on standard Arabic NLP tasks, while MARBERT \cite{abdul2021arbert} extended this capability to dialectal Arabic through large-scale pretraining on social media data.

For Algerian Dialect, DziriBERT \cite{abdaoui2021dziribert} introduced the first transformer model pretrained specifically on dialectal Algerian text, achieving state-of-the-art performance on several classification and sequence labeling tasks. Building upon this model, DziriBOT \cite{bechiri2026dziribot} proposed a task-oriented dialogue system for Algerian Dialect in the telecommunications domain, integrating intent classification, entity extraction, and rule-based dialogue management using the Rasa framework.

However, existing NLU systems remain predominantly text-based and assume clean textual input, making them vulnerable to error propagation when integrated with speech recognition components in real-world voice applications.

\subsection{Text-to-Speech for Arabic Dialects}

Neural text-to-speech systems have significantly advanced speech synthesis quality, particularly with architectures such as VITS and large-scale multilingual models. MMS-TTS \footnote{\url{https://huggingface.co/facebook/mms-tts}} introduced a massively multilingual speech framework covering over 1000 languages, including MSA, but with limited dialectal specialization.

To improve dialectal coverage, derivative systems such as vits-ar attempt to fine-tune MMS-based models for Arabic dialects. Similarly, Habibi-TTS \cite{chen2026habibi} introduced a unified framework for Arabic dialectal TTS, including Algerian Arabic, demonstrating promising results in multilingual settings.

Community-driven efforts have also explored Maghrebi dialects. For example, Dialect TTSv0.1-500M focuses on Moroccan Dialect using LoRA-based adaptation of large TTS backbones. While effective for Moroccan Arabic, these models do not generalize well to Algerian phonology and lexical structures. Other open-source systems such as Arabic-Speechsynthesis remain primarily trained on MSA data and do not explicitly address dialectal variation or code-switching.

Overall, TTS systems for Arabic dialects remain fragmented and under-resourced, with limited coverage of Algerian Dialect.

\subsection{Discussion and Research Gap}

Despite significant progress across ASR, NLU, and TTS, existing work remains largely modular and task-specific. Each component has been independently optimized, but there is a lack of unified systems that integrate all stages of spoken conversational pipelines.

ASR systems for Arabic dialects remain sensitive to limited data and are typically evaluated in isolation. NLU models achieve strong performance on text-based inputs but do not account for transcription noise introduced by speech recognition. TTS systems, while increasingly natural, are rarely integrated into end-to-end dialogue systems, limiting their evaluation in realistic conversational scenarios.

This fragmentation becomes particularly critical in low-resource settings such as Algerian Dialect, where data scarcity amplifies error propagation across components. To the best of our knowledge, no prior work has jointly addressed ASR, NLU, retrieval-augmented generation, and TTS within a unified speech-to-speech conversational framework for Algerian Dialect.

While commercial voice assistants provide end-to-end speech interfaces, they are not designed for dialectal Arabic and do not support Algerian Dialect in either recognition or generation.

To address this gap, we propose a complete end-to-end architecture that integrates all components into a single conversational pipeline, enabling consistent evaluation and realistic deployment in Algerian dialectal speech environments.

\section{Proposed End-to-End Architecture}
\begin{figure}[h!]
\centering
\includegraphics[width=\textwidth]{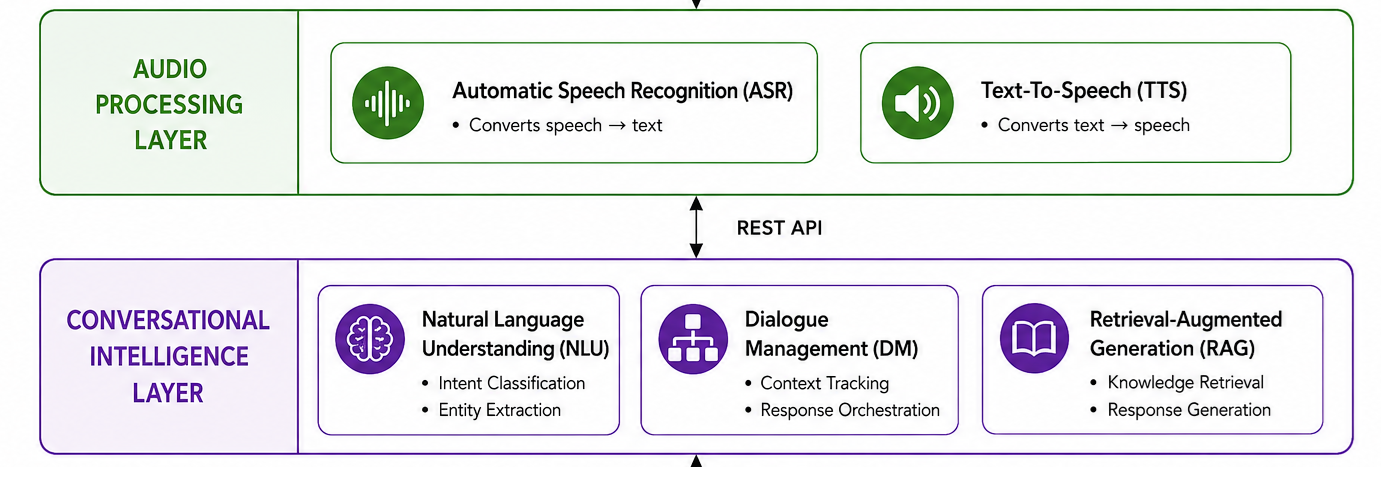}
\caption{Overview of the proposed DziriVoice end-to-end architecture.}
\label{fig:architecture_systeme}
\end{figure}

To address the challenges of speech-based interaction in Algerian Dialect, we propose a modular end-to-end conversational architecture that integrates Automatic Speech Recognition (ASR), Natural Language Understanding (NLU), Dialogue Management (DM), Retrieval-Augmented Generation (RAG), and Text-to-Speech (TTS) synthesis within a unified pipeline. The architecture follows a layered design principle that promotes component independence, scalability, and maintainability while enabling seamless speech-to-speech interaction. An overview of the proposed system is illustrated in Figure~\ref{fig:architecture_systeme}.

The architecture is organized into two major functional layers. (i) The \textit{Audio Processing Layer} is responsible for converting spoken input into textual representations and synthesizing spoken responses. (ii) The \textit{Conversational Intelligence Layer} performs semantic interpretation, dialogue orchestration, knowledge retrieval, and response generation.

The processing pipeline begins with the ASR module, which transcribes user speech into text. The resulting transcription is forwarded to the NLU component for intent classification and entity extraction. The Dialogue Manager then determines the appropriate action based on the detected intent, dialogue state, and conversational context. For queries covered by predefined business logic, responses are generated directly through the task-oriented dialogue system. Otherwise, requests are routed to a Retrieval-Augmented Generation module that combines semantic retrieval and large language model reasoning. Finally, generated responses are converted into natural Algerian Dialect speech through the TTS subsystem.

\subsection{Automatic Speech Recognition}

The ASR component constitutes the entry point of the conversational pipeline and is responsible for transforming spoken Algerian Dialect into text. Given the limited availability of dialectal speech resources, we investigate the adaptation of large-scale multilingual speech models through supervised fine-tuning on a domain-specific corpus collected for this work.

Two state-of-the-art architectures were evaluated: Whisper-medium and Wav2Vec2-XLS-R. Both models were fine-tuned using identical training and evaluation protocols to ensure a fair comparison. Whisper was selected as the primary deployment model due to its superior robustness to code-switching, spontaneous speech, and acoustic variability commonly observed in Algerian Dialect.

\subsection{Natural Language Understanding and Dialogue Management}
The NLU subsystem is implemented using the Rasa framework and is designed to extract semantic information from ASR transcriptions. The pipeline combines transformer-based contextual representations obtained from DziriBERT with traditional task-oriented dialogue components. Intent classification identifies the user's objective, while entity extraction captures relevant parameters required for task completion. \cite{bechiri2026dziribot}

Dialogue management follows a hybrid approach that combines rule-based reasoning with learned dialogue policies. Deterministic rules are used for critical business processes requiring predictable behavior, while machine-learned dialogue stories provide flexibility for multi-turn interactions. This hybrid design allows the system to maintain conversational consistency while adapting to diverse user behaviors.

To support realistic customer-service scenarios, the dialogue layer incorporates custom actions implementing business logic, contextual state tracking, form-based information collection, and recovery mechanisms for incomplete or ambiguous requests.

\subsection{Retrieval-Augmented Generation}

Although task-oriented dialogue systems provide reliable responses for predefined intents, they remain limited when confronted with open-domain or previously unseen questions. To address this limitation, we introduce a conditional Retrieval-Augmented Generation (RAG) module that acts as an intelligent fallback mechanism.

The retrieval stage relies on a domain-specific knowledge base containing structured information about telecommunications products, services, and offers. Documents are embedded using the multilingual \textit{paraphrase-multilingual-MiniLM-L12-v2} model and indexed using FAISS for efficient semantic search. To improve retrieval quality in dialectal settings, the system incorporates Dialect-specific normalization rules, category-aware filtering, and keyword-based reranking strategies.

Retrieved passages are provided as contextual evidence to a locally deployed Llama 3.2 3B Instruct model. The language model is constrained through prompt engineering techniques that enforce concise responses, restrict generation to retrieved evidence, and minimize hallucinations. This design enables the system to answer informational queries while preserving factual consistency and domain relevance.

\subsection{Text-to-Speech Synthesis}

The final stage of the pipeline converts textual responses into natural speech. Given the absence of publicly available TTS systems specifically designed for Algerian Dialect in the telecommunications domain, we constructed a dedicated speech corpus and investigated the adaptation of multilingual neural TTS architectures.

Two complementary approaches were explored. The first relies on XTTS-v2 enhanced through Low-Rank Adaptation (LoRA), enabling efficient specialization with limited training data. The second is based on VITS-ar, a VITS-derived architecture adapted for Arabic speech synthesis. These models were fine-tuned on a newly collected Algerian Dialect corpus to capture dialect-specific pronunciation patterns and code-switching phenomena.

By integrating ASR, NLU, DM, RAG, and TTS within a single framework, the proposed architecture provides a complete speech-to-speech conversational system tailored to the challenges of Algerian Dialect. The modular design allows each component to evolve independently while maintaining interoperability across the entire pipeline, facilitating future improvements and deployment in real-world customer-support environments.

\section{Experiments and Results}

\subsection{Speech Dataset Construction}
Due to the absence of publicly available speech corpora for Algerian Darija in the telecommunications domain, we constructed a dedicated multi-speaker speech dataset specifically designed for Automatic Speech Recognition (ASR) training and evaluation.

\paragraph{Data Collection.}
Text prompts were derived from real-world customer service interactions and cover 70 telecommunications-related intents, including balance inquiries, roaming services, subscriptions, customer support, and greetings. The corpus consists of 4,103 manually validated utterances written primarily in Arabic script, while preserving naturally occurring French code-switching and numeric substitutions commonly observed in Algerian Darija.

To facilitate large-scale data collection, we developed a custom web-based recording platform that allows contributors to record utterances directly through their browser. Each recording is automatically associated with its transcription, intent label, and speaker identifier. Audio files are standardized to 16 kHz mono format and stored together with their metadata in a structured corpus suitable for machine learning workflows.

\paragraph{Corpus Diversity.}
Rather than relying on artificial signal transformations during collection, diversity was introduced naturally through speaker and linguistic variability. The dataset was recorded by 14 Algerian speakers with different genders, vocal characteristics, speaking rates, and prosodic patterns. In addition, multiple alternative formulations were created for each intent, capturing lexical, syntactic, and morphological variations commonly encountered in real customer interactions.

This combination of speaker diversity and linguistic variation substantially increases corpus coverage and improves robustness to the variability inherent in spontaneous dialectal speech.

\paragraph{Dataset Statistics.}
The final corpus contains 4,103 utterances totaling approximately 2.68 hours of speech. Recordings range from 0.4 to 6.9 seconds in duration and cover 70 telecommunications intents. The corpus represents a realistic low-resource setting while preserving the linguistic characteristics of Algerian Darija, including frequent Arabic--French code-switching. More details in table \ref{tab:asr_dataset}.

Although the speaker distribution is naturally imbalanced, reflecting real-world participation rates, the resulting variability provides valuable acoustic diversity for model adaptation. The two most active speakers account for approximately 74\% of the recordings, while the remaining data are distributed across twelve additional speakers.

\begin{table}[h]
\centering
\caption{Statistics of the proposed ASR corpus.}
\label{tab:asr_dataset}
\begin{tabular}{lc}
\hline
Metric & Value \\
\hline
Utterances & 4,103 \\
Duration & 2.68 h \\
Speakers & 14 \\
Intents & 70 \\
Language & Algerian Darija + French CS \\
Sampling Rate & 16 kHz \\
Utterance Length & 0.4--6.9 s \\
\hline
\end{tabular}
\end{table}

\subsection{Automatic Speech Recognition (ASR)}

To identify the most suitable speech recognition architecture for Algerian Darija, we fine-tuned and evaluated two state-of-the-art pretrained models: Whisper-medium and Wav2Vec2-XLS-R. Both models were trained and evaluated on the same corpus using identical train, validation, and test partitions to ensure a fair comparison.

\paragraph{Data Preparation.}
The collected corpus was randomly partitioned into training (80\%), validation (10\%), and test (10\%) subsets using a fixed random seed ($42$) to ensure reproducibility. Each sample consists of an audio recording, its transcription, speaker identifier, intent label, and duration metadata. Audio files were standardized to 16~kHz mono format and organized into a unified structure compatible with the Hugging Face \texttt{Datasets} framework.

\subsubsection{Wav2Vec2-XLS-R Fine-Tuning}
We fine-tuned the multilingual Wav2Vec2-XLS-R model using a character-level tokenizer constructed directly from the training transcriptions. Text normalization was applied to reduce orthographic variability in both Arabic and Latin scripts, including character normalization and whitespace standardization.

To improve robustness, we employed on-the-fly audio augmentation during training through additive Gaussian noise and volume perturbation. In addition, the training set was enriched with the CAFE-small corpus \cite{lachemat2024cafe}, a spontaneous Algerian code-switching speech dataset, after applying the same preprocessing and normalization pipeline. Preliminary experiments with larger external corpora revealed substantial performance degradation caused by domain mismatch, confirming that domain relevance outweighs data quantity in specialized telecom ASR tasks.

Training was performed using the Connectionist Temporal Classification (CTC) objective. To accommodate hardware constraints, mixed-precision training, gradient checkpointing, gradient accumulation, and audio-length filtering were employed. Detailed hyper-parameters are given in table \ref{tab:wav2vec_hyperparams}.

\begin{table}[ht]
\centering
\caption{Fine-tuning Hyperparameters}
\label{tab:wav2vec_hyperparams}
\begin{tabular}{ll}
\toprule
\textbf{Hyperparameter} & \textbf{Value} \\
\midrule
Base model & facebook/wav2vec2-xls-r-300m \\
Training epochs & 20 \\
Batch size (per GPU) & 2--4 \\
Gradient accumulation steps & 16 \\
Effective batch size & 32 \\
Learning rate & $1 \times 10^{-4}$ \\
Warmup steps & 500 \\
Weight decay & 0.005 \\
Optimizer & AdamW \\
Mixed precision & FP16 \\
Gradient checkpointing & Enabled \\
Early stopping patience & 5 evaluation steps \\
Evaluation metric & Word Error Rate (WER) \\
Best model selection & Minimum WER \\
\bottomrule
\end{tabular}
\end{table}

\subsubsection{Whisper Fine-Tuning}
For Whisper, audio recordings were converted into log-Mel spectrograms using the official Whisper processor, while transcriptions underwent the same text normalization procedure adopted for Wav2Vec2.

Training data augmentation relied exclusively on signal transformations, including additive Gaussian noise and temporal stretching, generating multiple acoustic variations of each training sample. Unlike Wav2Vec2, experiments involving external corpora consistently degraded performance, with the inclusion of the CASABLANCA \cite{talafha2024casablanca} dataset increasing the Word Error Rate (WER) significantly. This observation further highlights the sensitivity of sequence-to-sequence ASR models to domain mismatch in low-resource settings.

Fine-tuning was conducted in two successive stages with progressively decreasing learning rates. The first stage focused on adapting the pretrained model to the acoustic and linguistic characteristics of Algerian Darija, while the second stage refined the learned representations to improve generalization and recognition accuracy. Detailed hyper-parameters are given in table \ref{tab:whisper_hyperparams}.

\begin{table}[ht]
\centering
\caption{Two-Stage Whisper Fine-Tuning Hyperparameters}
\label{tab:whisper_hyperparams}
\begin{tabular}{lcc}
\toprule
\textbf{Hyperparameter} & \textbf{Stage 1 (General)} & \textbf{Stage 2 (Polishing)} \\
\midrule
Learning rate & $1 \times 10^{-6}$ & $5 \times 10^{-7}$ \\
Warmup steps & 200 & 100 \\
Weight decay & 0.03 & 0.05 \\
Number of epochs & 5 & 3 \\
Batch size (per GPU) & 4 & 4 \\
Gradient accumulation & 4 steps & 4 steps \\
Effective batch size & 16 & 16 \\
Patience (Early Stopping) & 2 evaluations & 2 evaluations \\
FP16 / Gradient Checkpointing & Enabled & Enabled \\
\bottomrule
\end{tabular}
\end{table}

\subsection{Natural Language Understanding and Retrieval-Augmented Generation}
The Natural Language Understanding (NLU), Dialogue Management (DM), and Retrieval-Augmented Generation (RAG) components employed in this work are based on DziriBOT \cite{bechiri2026dziribot}, the first task-oriented chatbot developed for Algerian Darija in the telecommunications domain.

The present work can be viewed as a direct extension of DziriBOT from a text-based conversational system to a complete speech-to-speech voice assistant. Consequently, the conversational intelligence layer, including intent classification, entity extraction, dialogue management, and retrieval-augmented generation, was retained and integrated into the proposed architecture.

More specifically, the NLU component relies on a Rasa-based pipeline incorporating DziriBERT for intent classification and entity extraction, while dialogue management combines RulePolicy and TEDPolicy to support both deterministic and learned conversational flows. To handle user requests beyond predefined intents, a conditional Retrieval-Augmented Generation (RAG) module is employed, combining FAISS-based semantic retrieval with a locally deployed Llama 3.2 3B language model.

As the design and evaluation of these components have been extensively reported in DziriBOT \cite{bechiri2026dziribot}, they are not re-evaluated independently in this paper. Instead, they serve as the conversational backbone of the proposed system, allowing us to focus on the integration of speech recognition and speech synthesis technologies for end-to-end voice interaction in Algerian Darija.

\subsection{Text-to-Speech (TTS)}
To complete the speech-to-speech pipeline, we address the limited availability of Algerian Darija text-to-speech systems by constructing a dedicated telecom-oriented speech corpus and adapting two multilingual neural TTS architectures: XTTS-v2 and VITS-ar.

\subsubsection{TTS Corpus Construction}
A dedicated speech synthesis corpus was collected specifically for the telecommunications domain. Text prompts were derived from realistic customer-service interactions and written in Algerian Darija while preserving naturally occurring French code-switching. Recordings were produced by a native speaker and manually verified to ensure transcription quality and pronunciation consistency.

The resulting corpus contains approximately 50.7 minutes of speech and follows an LJSpeech-style metadata format linking each audio file to its corresponding transcription. Audio recordings were captured at 44.1~kHz and subsequently partitioned into training (95\%) and evaluation (5\%) subsets. Table \ref{tab:asr_dataset} summarizes the main characteristics of the corpus.

\subsubsection{XTTS-v2 Adaptation}
XTTS-v2 is a multilingual speech synthesis architecture combining a GPT-based autoregressive acoustic model with a HiFi-GAN neural vocoder. Its multilingual capabilities and support for speaker conditioning make it a suitable candidate for low-resource dialect adaptation.

To efficiently adapt the model to Algerian Darija, we employed Parameter-Efficient Fine-Tuning (PEFT) using Low-Rank Adaptation (LoRA). Only a small subset of parameters was updated while keeping the original pretrained model frozen. This strategy substantially reduces computational requirements while preserving the multilingual knowledge acquired during large-scale pretraining.

\subsubsection{VITS-ar Adaptation}
As a second approach, we fine-tuned VITS-ar, a multilingual Arabic speech synthesis model derived from MMS-TTS-ara. Unlike XTTS-v2, VITS performs end-to-end waveform generation and has demonstrated strong performance across several Arabic dialects.

Given the limited size of the available corpus, we adopted a decoder-only fine-tuning strategy, freezing the remaining network components and updating only the speech generation layers. This approach enables efficient adaptation to Algerian Darija while reducing the risk of overfitting.

The two architectures represent complementary adaptation paradigms: XTTS-v2 leverages parameter-efficient transfer learning through LoRA, whereas VITS-ar relies on partial model fine-tuning. Their comparative evaluation provides insights into the suitability of modern multilingual TTS architectures for low-resource Algerian dialect speech synthesis.

\section{Analysis and Discussion}
\subsection{ASR Results and Discussion}
We evaluated two state-of-the-art ASR architectures, Wav2Vec2-XLS-R-300M and Whisper-medium, on the held-out test set containing 411 utterances. Table~\ref{tab:asr_results} summarizes the obtained Word Error Rates (WER).

\begin{table}[ht]
\centering
\caption{ASR performance on the Algerian Darija test set.}
\label{tab:asr_results}
\begin{tabular}{lcc}
\toprule
Model & Training Data & WER (\%) \\
\midrule
Wav2Vec2-XLS-R-300M & telecom & 31.67 \\
Wav2Vec2-XLS-R-300M & telecom + CAFE & 29.04 \\
Whisper-medium & telecom & 15.40 \\
Whisper-medium & telecom + Augmentation & \textbf{13.74} \\
Whisper-medium & telecom + Casablanca & 36.46 \\
\bottomrule
\end{tabular}
\end{table}

Despite the limited size of the training corpus (2.68 hours), both architectures benefited substantially from transfer learning. The best-performing configuration, Whisper-medium trained with signal-based augmentation, achieved a WER of 13.74\%, outperforming the strongest Wav2Vec2 configuration by approximately 53\% relative. To the best of our knowledge, this is the first reported ASR result for Algerian Darija in the telecommunications domain.

The superiority of Whisper can be attributed to three factors. First, its encoder-decoder architecture incorporates an implicit language model that better captures contextual information and code-switching patterns between Darija and French. Second, Whisper benefits from significantly larger multilingual pretraining data, including extensive Arabic speech resources. Third, its sequence-to-sequence formulation enables stronger contextual decoding compared to the frame-level CTC objective employed by Wav2Vec2.

An additional observation concerns domain adaptation. While the controlled inclusion of the CAFE corpus improved Wav2Vec2 performance, incorporating the larger CASABLANCA corpus into Whisper training resulted in a substantial degradation (36.46\% WER). This finding highlights the importance of domain relevance over dataset size in low-resource ASR settings.

\subsubsection{Error Analysis}
To better understand model behavior, we manually inspected transcription errors produced by both systems. Three main categories emerged:

\begin{itemize}
\item \textbf{Phonetic confusions}: substitutions involving acoustically similar sounds, particularly word-final vowels and dialect-specific pronunciations.
\item \textbf{Lexical substitutions}: replacement of a word by another semantically plausible expression, typically caused by language-model bias.
\item \textbf{Tokenization and word-boundary errors}: incorrect segmentation of morphologically complex Darija words.
\end{itemize}

Qualitative analysis revealed a fundamental difference between the two architectures. Whisper errors were predominantly phonetic and often preserved the semantic meaning of the utterance, allowing downstream NLU components to remain functional. In contrast, Wav2Vec2 generated more severe lexical substitutions capable of altering the intended meaning and causing intent classification failures.

These observations suggest that WER alone does not fully capture the practical impact of transcription errors in conversational systems. From an end-to-end perspective, semantically preserving errors are considerably less harmful than intent-altering substitutions.

\subsubsection{Discussion}
Based on both quantitative and qualitative evaluations, Whisper-medium was selected as the ASR component of the final voice assistant pipeline. Although it requires greater computational resources than Wav2Vec2, the substantial gain in recognition accuracy justifies this choice.

The current implementation exhibits an average inference latency of approximately 16 seconds per interaction on the available hardware. This limitation is primarily computational rather than architectural and can be mitigated through GPU deployment, model quantization, and optimized inference frameworks. Future work will investigate these directions while expanding the speech corpus to improve robustness across regional and demographic variations of Algerian Darija.

\subsection{TTS Results and Discussion}
We evaluated the two fine-tuned TTS architectures, XTTS-v2 with LoRA adaptation and VITS-ar with decoder-only fine-tuning, using both perceptual quality and inference efficiency metrics.

\subsubsection{Mean Opinion Score (MOS) Evaluation}
Speech quality was assessed through a subjective listening test involving 15 native speakers of Algerian Darija. Participants evaluated 30 synthesized utterances from each model using a five-point Mean Opinion Score (MOS) scale, resulting in a total of 450 ratings per model. The evaluation focused on speech naturalness and intelligibility.

Table~\ref{tab:mos_results} summarizes the obtained results.

\begin{table}[ht]
\centering
\caption{MOS evaluation results (mean $\pm$ standard deviation).}
\label{tab:mos_results}
\begin{tabular}{lccc}
\toprule
Model & Naturalness & Intelligibility & Overall MOS \\
\midrule
XTTS-v2 + LoRA & 4.12 $\pm$ 0.58 & 4.51 $\pm$ 0.45 & \textbf{4.31 $\pm$ 0.52} \\
VITS-ar (Decoder FT) & 2.70 $\pm$ 0.71 & 3.54 $\pm$ 0.62 & 3.12 $\pm$ 0.67 \\
\bottomrule
\end{tabular}
\end{table}

XTTS-v2 consistently outperformed VITS-ar across all evaluation dimensions, achieving an overall MOS of 4.31 compared to 3.12. The observed improvement of 1.19 MOS points demonstrates a substantial perceptual advantage in both naturalness and intelligibility. Furthermore, the lower variance observed for XTTS-v2 indicates stronger agreement among evaluators, suggesting more stable synthesis quality across different utterances.

\subsubsection{Inference Latency Analysis}
In addition to perceptual quality, we evaluated synthesis latency under several response lengths using the optimized XTTS-v2 configuration. Table~\ref{tab:tts_latency} reports synthesis time and Real-Time Factor (RTF).

\begin{table}[ht]
\centering
\caption{XTTS-v2 inference latency on CPU.}
\label{tab:tts_latency}
\begin{tabular}{lccc}
\toprule
Configuration & Synthesis Time (s) & Audio Duration (s) & RTF \\
\midrule
Baseline & 82.4 & 19.0 & 4.34 \\
Short response & 10.8 & 2.1 & 5.14 \\
Medium response & 17.5 & 3.5 & 5.00 \\
Long response & 21.9 & 4.9 & 4.47 \\
\bottomrule
\end{tabular}
\end{table}

The optimized configuration reduced synthesis time by approximately 73\% compared to the baseline setup. Nevertheless, the resulting RTF remains around 4.7, indicating that the system requires approximately 4.7 seconds of computation to generate one second of speech. While this performance is sufficient for prototype deployment, real-time operation would require GPU acceleration or additional model optimization.

\subsubsection{Discussion}
The superiority of XTTS-v2 can be attributed to both architectural and adaptation-related factors. First, its GPT-based autoregressive architecture better captures long-range prosodic dependencies and code-switching patterns frequently encountered in Algerian Darija. Second, the LoRA-based adaptation strategy enables efficient specialization while preserving the multilingual knowledge acquired during large-scale pretraining. In contrast, the decoder-only fine-tuning strategy adopted for VITS-ar appears insufficient to fully adapt the model to the phonetic and prosodic characteristics of Algerian Darija.

The obtained MOS score of 4.31 demonstrates that high-quality dialectal speech synthesis can be achieved with a relatively small corpus (50.7 minutes) when leveraging modern multilingual foundation models and parameter-efficient adaptation techniques. Based on both perceptual quality and overall robustness, XTTS-v2 was selected as the speech synthesis component of the final voice assistant pipeline.

\section{Conclusion}
This paper presented DZIRI VoiceBot, an end-to-end speech-to-speech conversational system for Algerian Dialect, a low-resource dialect that remains largely unsupported by existing voice technologies. To address the challenges of Algerian Dialect, we proposed a unified architecture integrating Automatic Speech Recognition, Natural Language Understanding, Retrieval-Augmented Generation, and Text-to-Speech synthesis. Beyond the system itself, this work contributes new linguistic resources for Algerian Dialect, including dedicated corpora for speech recognition, dialogue understanding, and speech synthesis in the telecommunications domain. These resources constitute an important step toward reducing the data scarcity that has historically limited research on Algerian dialectal speech technologies.

Experimental results demonstrate the effectiveness of the proposed approach across all stages of the conversational pipeline. Fine-tuning Whisper-medium achieved a Word Error Rate of 13.74\%, establishing a strong benchmark for Algerian Dialect speech recognition in a domain-specific setting. The hybrid Rasa--DziriBERT pipeline obtained 98.4\% intent classification accuracy and 93.9\% entity-level F1-score, while the conditional RAG module successfully extended system coverage beyond predefined intents. Furthermore, the adaptation of XTTS-v2 and VITS-ar demonstrated the feasibility of generating natural and intelligible Algerian Dialect speech despite limited training data.

More importantly, this work demonstrates that building a production-oriented voice assistant for a low-resource Arabic dialect is achievable through the combination of targeted data collection, transfer learning from multilingual foundation models, and modular system design. To the best of our knowledge, this is the first study to integrate and evaluate ASR, NLU, RAG, and TTS within a complete speech-to-speech conversational framework for Algerian Dialect.

Despite these encouraging results, several challenges remain. The limited size of the collected speech corpora, the computational cost of retrieval-augmented generation, and the restricted coverage of the knowledge base highlight opportunities for future improvements. Expanding dialectal datasets, improving real-time inference through model optimization, and incorporating larger conversational language models constitute promising directions for future work. More broadly, we hope that this work serves both as a practical foundation and as a methodological reference for future research on Algerian Dialect and, more generally, on under-resourced Arabic dialects.

\bibliographystyle{unsrtnat}
\bibliography{references} 
\end{document}